\documentclass[10pt, conference]{IEEEtran}
\usepackage{orcidlink}

\usepackage{amsmath,amssymb,amsfonts}
\usepackage{pifont}

\usepackage{amsthm}
\usepackage{xcolor}
\usepackage[figuresright]{rotating}

\usepackage{graphicx}
\graphicspath{{/}{fig/}}

\usepackage{array}
\usepackage{textcomp}
\usepackage{multirow}
\usepackage{booktabs}

\usepackage{mathtools}
\usepackage{breqn}
\usepackage{float}

\usepackage{pgfplots}
\pgfplotsset{compat=1.7}
\usepgfplotslibrary{groupplots}

\usepackage{caption}
\usepackage{subcaption}

\usepackage{tabularx}
\usepackage{hyperref}
\usepackage{flushend}
\usepackage{algorithmic}
\usepackage[vlined, ruled, shortend]{algorithm2e}

\usepackage{enumitem}
\usepackage{gensymb}

\newlength\figureheight
\newlength\figurewidth
\SetAlCapNameFnt{\footnotesize}
\SetAlCapFnt{\footnotesize}

\title{
    UAV Tracking with Solid-State Lidars: \\Dynamic Multi-Frequency Scan Integration 
}

\author{
    \IEEEauthorblockN{
        \vspace{1em}
        Iacopo Catalano\IEEEauthorrefmark{2}\,\orcidlink{0000-0000-0000-0000},
        Ha Sier\IEEEauthorrefmark{2}\,\orcidlink{0000-0000-0000-0000},
        Xianjia Yu\IEEEauthorrefmark{2}\,\orcidlink{0000-0002-9042-3730},
        Tomi Westerlund\IEEEauthorrefmark{2}\,\orcidlink{0000-0002-1793-2694},
        Jorge Peña Queralta\IEEEauthorrefmark{2}$^{,}$\IEEEauthorrefmark{3}\,\orcidlink{0000-0003-3091-3217}
    }
    \IEEEauthorblockA{
        \normalsize
        \IEEEauthorrefmark{2}\href{https://tiers.utu.fi}{Turku Intelligent Embedded and Robotic Systems (TIERS) Lab, University of Turku, Finland}.\\
        \IEEEauthorrefmark{3}\href{https://scai.ethz.ch/}{SCAI Laboratory at SPZ, Swiss Federal School of Technology in Zurich - ETH Zurich, Switzerland}.\\
        Emails: \textsuperscript{1}\{imcata, sierha, xianjia.yu, tovewe\}@utu.fi, \textsuperscript{1}\{jorge.penaqueralta\}@hest.ethz.ch\\[+6pt]
    }
}

\begin{document}

\maketitle
\thispagestyle{empty}
\pagestyle{empty}



\begin{abstract}%
    \label{sec:abstract}%
    With the increasing use of drones across various industries, the navigation and tracking of these unmanned aerial vehicles (UAVs) in challenging environments, namely GNSS-denied environments, have become critical issues. In this paper, we propose a novel method for a ground-based UAV tracking system using a solid-state LiDAR, which dynamically adjusts the LiDAR frame integration time based on the distance to the UAV and its speed. Our method fuses two simultaneous scan integration frequencies for high accuracy and persistent tracking, enabling reliable state estimation of the UAV even in challenging scenarios. The application of the Inverse Covariance Intersection method and Kalman filters allows for better tracking accuracy and can handle challenging tracking scenarios. Compared to previous works in solid-state lidar tracking, this paper presents a more complete and robust solution.

    We have performed a number of experiments to evaluate the performance of the proposed tracking system and identify its limitations. Our experimental results demonstrate that the proposed method clearly outperforms the baseline method and ensures tracking is more robust across different types of trajectories.

    The code is open-source and available at \href{https://github.com/TIERS/dynamic\_scan\_tracking}{github.com/TIERS/ dynamic\_scan\_tracking}.
\end{abstract}

\begin{IEEEkeywords}

    UAV; Tracking; Solid-State LiDAR; Multi-Scan Integration; Adaptive Scanning

\end{IEEEkeywords}
\IEEEpeerreviewmaketitle


\section{Introduction}\label{sec:introduction}

Unmanned Aerial Vehicles (UAVs) are becoming more prevalent in various application domains due to their mobility and ease of deployment~\cite{tsouros2019review, wang2019surveying}. Equipped only with a flight controller and a basic sensor suite, they are ideal as mobile and easily deployable sensing platforms~\cite{gawel2018aerial, nex2014uav}. In recent years, researchers have been focusing on the navigation of UAVs in GNSS-denied environments~\cite{nieuwenhuisen2016autonomous, sier2023uav, jiang2021anti} as well as state estimation in both single and multi-UAV systems~\cite{queralta2020collaborative, queralta2020autosos}.

The use of UAVs is becoming also more prevalent in multi-robot systems, where tracking between robots is a common aspect of relative or global state estimation methods~\cite{queralta2022vio, bai2020cooperative}. From the perspective of deployment within multi-robot systems, being able to track UAVs from an unmanned ground vehicle (UGV) enables miniaturization and higher degrees of flexibility lowering the need for high-accuracy onboard localization~\cite{petrlik2020robust, rouvcek2019darpa}. This is because the UGV can act as a base station, providing the UAV with necessary data and allowing it to operate in areas where GNSS signals may not be available.

\begin{figure}
    \centering
    \includegraphics[width=.49\textwidth]{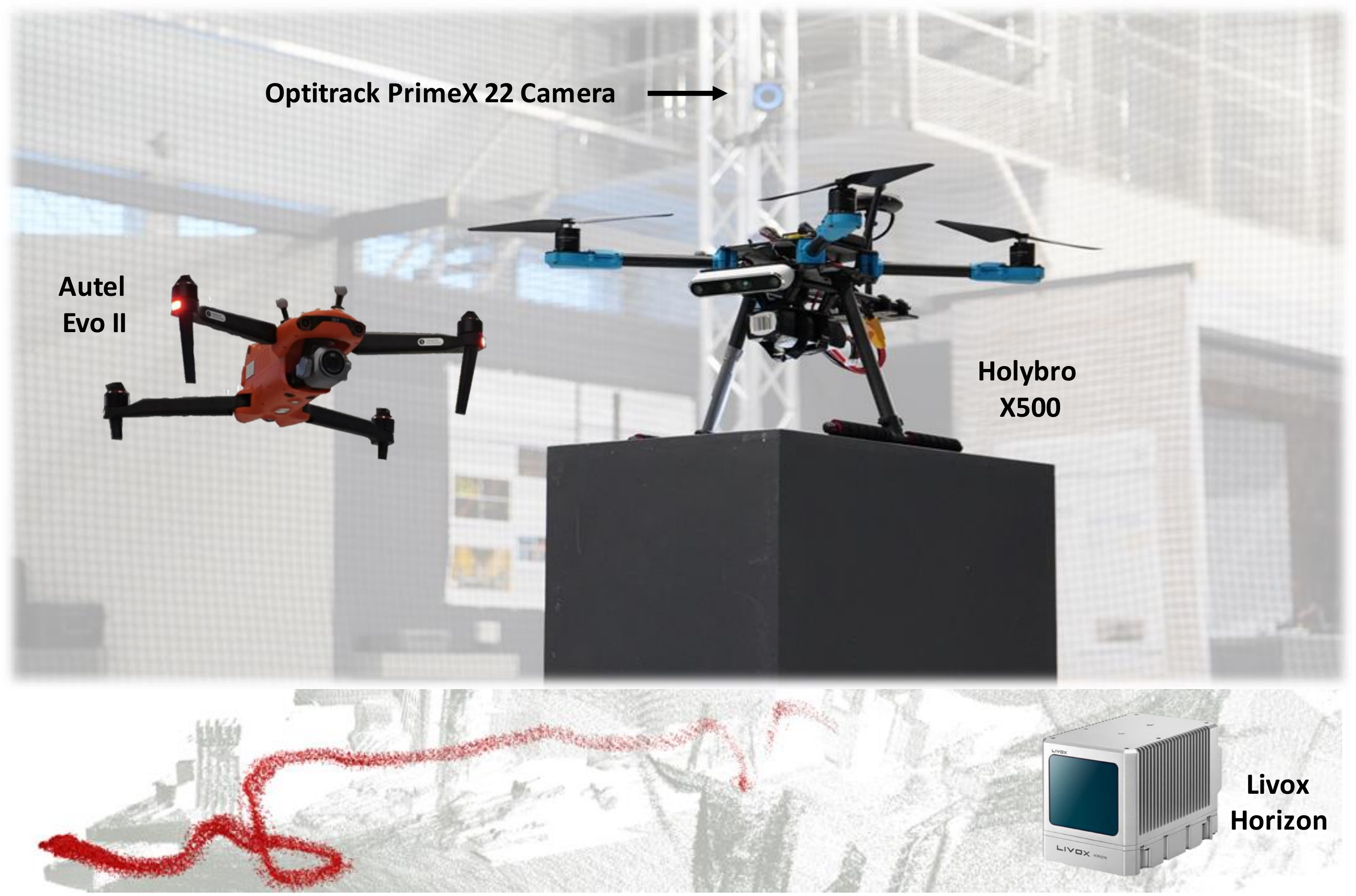}
    \caption{Illustration of hardware used in the experiments.}
    \label{fig:concept}
\end{figure}

Despite the significant progress made in UAV tracking using GNSS and other sensors, there are still limitations and challenges to be addressed: GNSS signals may not be available in certain areas, such as indoor environments or urban canyons, which limits the accuracy and reliability of UAV tracking. Furthermore, existing methods may rely on expensive hardware or require high levels of computational power, which limits their practicality and scalability~\cite{li2020towards, osco2021review}. These limitations and challenges motivate the demand for new approaches that are more robust, accurate, and efficient, and that can operate in GNSS-denied environments. In addition to the above, external tracking of UAVs from the ground has also gained importance within the context of counter-UAV solutions~\cite{samaras2019deep}.

Solid-state LiDARs are a recent development in long-range scanning technology that produce high-density point clouds, making them ideal for tracking objects in three-dimensional space, such as UAVs~\cite{li2020towards}. With non-repetitive scan patterns, they can generate dense point clouds with adjustable frequencies and varying field of view (FoV) coverage. In this article, we build upon our previous works~\cite{qingqing2021adaptive, catalano2023evaluating} to provide a more sophisticated approach that enables real-time UAV tracking. Specifically, we design and develop an algorithm to dynamically adjust the scan integration time, or frequency, based on the tracking state (position and velocity). We do this in parallel for two different integration frequencies, both dynamically adjusted. A higher frequency, as preliminary results in~\cite{li2020towards} show, allows for more accurate tracking, while the lower frequency enables persistence over time. Importantly, we address here two main limitations in~\cite{li2020towards}. First, the lack of tracking, as the previous algorithm simply tracks a UAV by iterative detections. Second, while the concept is similar, the integration in~\cite{li2020towards} is done offline and in post-processing. We also provide a quantifiable comparison with ground truth data, extending the benchmarks in~\cite{catalano2023evaluating} with the new dynamic multi-frequency integration approach. Finally, we no longer require a priori knowledge of the UAV shape but only its approximate size.

In summary, our main contributions with respect to our previous work are in both data processing (real-time adaptive dynamic integration) and a tracking algorithm:

\begin{enumerate}[label = (\roman*)]
    \item an algorithm that adjusts the LiDAR frame integration time, or frequency, dynamically based on the UAV speed and distance from the sensor. We integrate consecutive scans in a sliding window manner to retain the most recent information about the state of the UAV.
    \item a novel dual tracking approach using a Kalman filter variant that combines the two scan integration frequencies into one single state estimation using inverse covariance intersection. We evaluate the tracking performance during the experiments with ground truth data generated by a motion capture system.
\end{enumerate}

Moreover, we provide a method to detect the initial position of the UAV using the object detection algorithm YOLOV5 over range images generated from a solid-state LiDAR point cloud.

In the following sections, we will first provide a brief review of related work on UAV tracking and LiDAR-based sensing in Section II. Then, we will describe the proposed method in detail, including the dynamic adjustment of LiDAR frame integration time and the use of Kalman filters in Section III. We will present experimental results demonstrating the accuracy and robustness of the proposed method in tracking UAVs in Section V, where we will follow by providing a discussion on the tracking initialization method to finally conclude in Section VI with future directions for research.


\section{Related Work} \label{sec:related_work}

LiDAR systems are often employed for detecting and tracking objects, including UAVs. However, tracking UAVs with LiDAR can be challenging due to their small size, varied shapes and materials, high speed, and unpredictable movements.

In order to overcome these challenges, researchers have explored various methods to overcome the limitations of 3D LiDAR technology and improve the detection and tracking of UAVs. One approach involves conducting a probabilistic analysis of detections using a LiDAR mounted on a rotating turret, as described in~\cite{dogru2022drone}. The rotating turret enables a wider field of view coverage with fewer LiDAR beams while continuously tracking only a small number of hits.

A different strategy involves combining a segmentation approach and a simple object model while leveraging temporal information, as demonstrated in~\cite{razlaw2019detection}. This approach has been shown to reduce the parametrization effort and generalize well in different settings. Alternatively, ~\cite{wang2021study} perform the detection of UAV using Euclidean distance clustering and a particle filter algorithm to complete tracking. Overall, the use of LiDAR technology offers various methods for improving the detection and tracking of UAVs, and researchers are continually exploring new techniques to overcome the unique challenges posed by these small, fast-moving, and unpredictable objects.

Deploying UAVs from a ground robot requires careful consideration of the relative localization between different devices. To address this challenge, Li et al.~\cite{qingqing2021adaptive} proposed a multi-modal approach that combines three tracking modalities and integrates multiple scans to adjust the density and size of the point cloud that needs to be processed. Similarly, in our previous work~\cite{catalano2023evaluating} we evaluate the effectiveness of the multi-scan integration technique and introduce a Kalman filter to perform tracking as a replacement for the previous track-after-detect approach.

Furthermore, in \cite{pritzl2022cooperative}, a cooperative navigation framework is proposed to address the challenge of guiding a secondary UAV with unreliable self-localization employing a primary UAV equipped with a LiDAR sensor. The primary UAV's detector estimates an occupancy voxel map, while the tracker module compensates for detection delay and improves position estimation through a Kalman Filter-based multi-target tracking technique.

In a different approach, Sier et al~\cite{sier2023uav} adopt the LiDAR-as-a-camera concept fusing images and point cloud data generated by a single LiDAR sensor to track UAVs without a priori knowledge. Employing a custom YOLOv5 model trained on panoramic images, they are capable of bringing computer vision capabilities on top of the LiDAR itself.

Additionally, Kim et al \cite{kim2017lidar} conducted a study on autonomous UAV landing on a moving ground vehicle. By employing a clustering algorithm, the UAV is identified within the point cloud data, allowing for position estimation. Moreover, the method eliminates outliers, such as trees or people, by configuring a region of interest in close proximity to the UAV.

Another technique, departing from the typical sequence of track-after-detect, is to leverage motion information by searching for minor 3D details in the 360$^{\circ}$ LiDAR scans of the scene and analyzing the trajectory of the tracked object to classify UAVs and non-UAV objects by identifying typical movement patterns~\cite{hammer2018potential, hammer2018lidar}.
 

\begin{figure*}
    \centering
    \includegraphics[width=\textwidth]{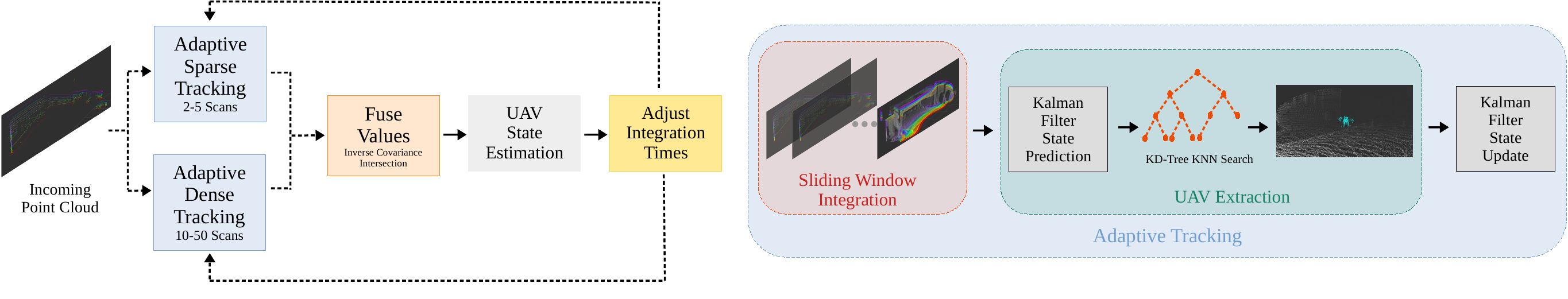}
    \caption{Overview of the proposed method: tracking is simultaneously performed at two different integration time ranges and then fused using the Inverse Covariance Intersection (ICI) method. The fused estimate is used to refine the estimations and integration times of both estimators.}
    \label{fig:system_overview}
\end{figure*}

\section{Methodology}

\subsection{Baseline Method}

To establish a baseline for comparison, we investigate a tracking approach that keeps the integration time constant along the UAV trajectory. Unlike our proposed method, the approach in~\cite{catalano2023evaluating} performs tracking without using a sliding window. Our baseline method uses a sliding window to integrate the LiDAR scans and capture more relevant information while keeping the integration time constant.

Overall, the baseline method is similar to our adaptive method, with the main difference being that in the adaptive method, we vary the integration time based on the UAV motion dynamics. In the following section, we provide further analysis of the results and compare the performance of the adaptive method to that of the baseline.

\subsection{Formulation}

We propose two simultaneous tracking estimators on two different scan frequencies running in parallel where the integration time $I$, defining the number of scans to accumulate, is dynamically adjusted to optimize the point cloud density as depicted in Fig.~\ref{fig:system_overview}:

\begin{enumerate}[label=(\roman*)]
    \item Adaptive Sparse Tracking (AST): Sparse point clouds are integrated up to 5 consecutive scans. This process provides high accuracy for estimating the UAV position and speed, but only tracks it through a reduced number of points and not necessarily in all frames.
    \item Adaptive Dense Tracking (ADT): The number of scans ranges between 10 and 50. The extracted point cloud representing the UAV is distorted by motion, which reduces the accuracy of localization and speed estimation, but the tracking is more robust and persistent as the UAV can be recognized in most frames.
\end{enumerate}

In the following formulation, we denote discrete steps as $k$ due to the discrete nature of the set of consecutive point clouds.

Let $\mathcal{P}^{}_k(I_{r}^k) = \{ \mathbf{p}_{1}^{k}, \mathbf{p}_{2}^{k}, \dots, \mathbf{p}_{n_k}^{k} \}$ be the set of $n_k$ points in the point cloud generated by the LiDAR sensor at time step $k$ using an integration time $I_{r}^k$, where $r$ is the range of the interval. The objective of the tracking algorithm is to identify the subset of points in $\mathcal{P}^{}_k(I_{r}^k)$ that corresponds to the UAV, denoted $\mathcal{P}^{k}_{\text{UAV}}$, to estimate its position and velocity, and consequently adjust the integration time for the next point cloud in the two integration time intervals, $I^{k}_{\text{AST}}, I^{k}_{\text{ADT}}$.

To initiate the tracking process, we assume the initial position of the UAV to be known. The point cloud $\mathcal{P}^{}_k(I_{r}^k)$ is integrated by accumulating the number of scans defined by $I_{r}^k$ in a sliding window fashion. This allows us to retain the most recent information about the state of the UAV in the point cloud. We then employ a Nearest-Neighbor Search (NNS) algorithm to identify the points in the point cloud that are closest to the predicted position of the UAV, based on its initial position. To improve the reliability and accuracy of the tracking results, we leverage a priori information about the dimensions of the tracked object. Specifically, the NNS is constrained to a search radius of $r$, which is set to half of the largest dimension (length, width or height) of the UAV. This allows us to constrain the NNS to a smaller volume around the estimated position, leading to faster and more accurate search results.

Next, we estimate a new position for the UAV by averaging the extracted points, which serves as the measurement in the Extended Kalman Filter (EKF) update step. To account for the large distances between scans caused by the velocity of the UAV, we have chosen to prioritize the most recent point clouds. This means that points in these more recent clouds are given greater importance than those in earlier ones. We accomplish this by assigning a weight to each point based on its timestamp $t_{p}^{}$, which follows the formula shown in Equation~\eqref{eq:points_weight}:

\begin{equation}\label{eq:points_weight}
   w^{}_{\text{p}} = \exp\left[- \gamma \times \left(t_{\text{scan}}^{} - t_{\text{p}}^{} \right)\right]
\end{equation}

where $t_{\text{scan}}^{}$ corresponds to the time at which the latest scan is acquired.

For the prediction step of the EKF, we use a Constant Turn Rate and Velocity (CTRV) motion model, commonly used for airborne tracking systems~\cite{blackman1999design}. In our case, we extended this model to the 3D scenario by incorporating a Constant Velocity (CV) motion model for the $z$ coordinate. We opted for this model for its proven robustness in the literature \cite{montanez2023application, guo20223d}.

The state space

\begin{equation}\label{eq:ctrv_state_space}
    \mathbf{x}=[\ \ x\quad y\quad z\quad v\quad \psi\quad \dot{\psi}\ ]^{T}
\end{equation}

can be transformed by the non-linear state transition.

\begin{equation}\label{eq:ctrv_state_transition}
    \mathbf{x}^{}_{k+1} = \mathbf{x}^{}_k + 
    \begin{bmatrix}
        \frac{v^{}_k}{\dot{\psi}^{}_k} \left(\sin\left(\psi^{}_k + \dot{\psi}^{}_k \Delta t\right) - \sin\left(\psi^{}_k\right)\right) \cr 
        \frac{v^{}_k}{\dot{\psi}^{}_k} \left(- \cos\left(\psi^{}_k + \dot{\psi}^{}_k \Delta t\right) + \cos\left(\psi^{}_k\right)\right) \cr 
        \dot{z} \Delta t \cr
        0 \cr
        \dot{\psi}^{}_k \Delta t \cr
        0
    \end{bmatrix}
\end{equation}

Next, the updated predicted position is used as input to the NNS algorithm to obtain an updated measurement. This measurement is then used in the next iteration of the EKF to further improve the accuracy of the predicted position. For more details, we refer the reader to Algorithm~\ref{alg:tracking}.

\begin{algorithm}[t]
    \footnotesize
	\caption{\footnotesize UAV tracking with adaptive scan integration}
	\label{alg:tracking}
	\KwIn{\\ 
    Adaptive Sparse and Dense Tracking int. rates:  \hfill $\left\{I^{k-1}_{\text{AST}}, I^{k-1}_{\text{ADT}}\right\}$ \\ \vspace{.42em}
    3D lidar point clouds: \hfill $\left\{\mathcal{P}_{k}\left(I_{\text{AST}}^{k-1}\right), \mathcal{P}_{k}\left(I_{\text{ADT}}^{k-1}\right)\:\right\}$ \\ \vspace{.42em}
    Last known UAV state: \hfill $ \left(\textbf{x}^{k-1}_{\text{UAV}}, {\dot{\textbf{x}}_{\text{UAV}}}^{k-1}\right)$ \\ \vspace{.42em}
	}
	\KwOut{\\
	    \begin{tabular}{ll}
	        UAV state:      & \{${\textbf{x}^{k}_{\text{UAV}}}, \:\dot{\textbf{x}}^{k}_{\text{UAV}}\}$ \\[+0.2em]
	    \end{tabular}
	}  
	%
	\SetKwFunction{FSub}{$uav\_tracking\left(\mathcal{P}, \:I, \:{\textbf{x}^{k-1}_{\text{UAV}}}, \:\dot{\textbf{x}}^{k-1}_{\text{UAV}}\right)$}
    \SetKwProg{Fn}{Function}{:}{}
    \BlankLine
    \Fn{\FSub}
    {
            \vspace{.3em}
            \begin{tabular}{ll}
                UAV pos estimation:     & $\hat{\textbf{x}}_{\text{UAV}}^{k}, \dot{\textbf{x}}^{k}_{\text{UAV}} = \mathcal{KF}^{}_{\text{prediction}}\left(\textbf{x}^{k-1}_{\text{UAV}}, \dot{\textbf{x}}^{k-1}_{\text{UAV}}\right);$ \\[+0.4em]
                Generate KD Tree:       & $kdtree \leftarrow \mathcal{P};$ \\[+0.4em]
                UAV points:             & $\mathcal{P}^{k}_{\text{UAV}} = NNS\left(kdtree, \: \hat{\textbf{x}}_{\text{UAV}}^{k}\right);$ \\[+0.4em]
                UAV measurement:   & ${\textbf{z}^{k}_{\text{UAV}}} = \frac{1}{\lvert\mathcal{P}^{k}_{\text{UAV}}\rvert}\sum_{x\in\mathcal{P}^{k}_{\text{UAV}}} x;$\\[+0.8em]
                UAV state estimation:   & ${\textbf{x}^{k}_{\text{UAV}}} = \mathcal{KF}^{}_{\text{update}}\left(\textbf{z}^{k}_{\text{UAV}}\right);$
                \\
            \end{tabular}
            \KwRet ${\textbf{x}^{k}_{\text{UAV}}}, \dot{\textbf{x}}^{k}_{\text{UAV}};$
    }
	\BlankLine

    \tcp{Adaptive Sparse Tracking}
    \ForEach{new $\mathcal{P}_{k}\left(I^{k}_{\text{AST}}\right)$}{
        ${\textbf{x}^{k''}_{\text{UAV}}} =  uav\_tracking\left(\mathcal{P}_{k}\left(I^{k}_{\text{AST}}\right), I^{k}_{\text{AST}}, \textbf{x}^{k-1}_{\text{UAV}}, \dot{\textbf{x}}^{k-1}_{\text{UAV}}\right);$ \\
    }
    
    \tcp{Adaptive Dense Tracking}
    \ForEach{new $\mathcal{P}_{k}\left(I^{k}_{\text{ADT}}\right)$}{
        ${\textbf{x}^{k'}_{\text{UAV}}} =  uav\_tracking\left(\mathcal{P}_{k}\left(I^{k}_{\text{ADT}}\right), I^{k}_{\text{ADT}}, \textbf{x}^{k-1}_{\text{UAV}}, \dot{\textbf{x}}^{k-1}_{\text{UAV}}\right);$ \\
    }

    \tcp{Inverse Covariance Intersection}
    $\{\textbf{x}^{k}_{\text{UAV}}, \dot{\textbf{x}}^{k}_{\text{UAV}}\} \leftarrow new\_state\_estimation\left({\textbf{x}^{k'}_{\text{UAV}}}, {\textbf{x}^{k''}_{\text{UAV}}}\right);$\\[+0.4em]
    
    $\{I^{k}_{\text{AST}}, I^{k}_{\text{ADT}}\} \leftarrow adjust\_integration\_rates \left(\textbf{x}^{k}_{\text{UAV}}, \dot{\textbf{x}}^{k}_{\text{UAV}}\right);$ \\[+0.4em]

\end{algorithm}

Within each integration time interval, the integration time is adjusted based on the distance and velocity of the tracked object. To avoid motion blur, shorter integration times are used for closer and faster-moving objects, while longer integration times are used for more distant and slower-moving objects. This adjustment is based on Equation~\eqref{eq:adaptive_time}.

\begin{equation}\label{eq:adaptive_time}
    I_{r}^k = I_{\text{min}}^k + \left(I_{\text{max}}^k - I_{\text{min}}^k\right) \left( \frac{d}{d^{}_{\text{max}}} \right)
\end{equation}

Here, $I_{\text{max}}^k$ and $I_{\text{min}}^k$ define the upper and lower bound for the integration time within each interval, and $d$ represents the current distance of the UAV from the sensor. As for $d^{}_{\text{max}}$, it represents the maximum distance within an integration time interval for which a minimum of 4 points on the UAV were recognized in the point cloud $\mathcal{P}_k(I_{r}^k)$. This distance was empirically calculated by manually determining the number of points detected at a distance of 4\,m. Moreover, a simplified approximation of the number of points hitting the target was designed assuming that the point density remains constant, and the primary factor influencing the number of points is the spreading of points over distance due to the LiDAR's angular resolution as per Equation~\eqref{eq:num_points}:

\begin{equation}\label{eq:num_points}
   N^{} = \frac{S^{2}_{}}{d^2 \theta}
\end{equation}

where $S$ represents the side length of the target surface, $d$ is the distance from the sensor and $\theta$ represents the LiDAR's angular resolution in radians which accounts for the angular spacing between individual points.

The maximum distance $d^{}_{\text{max}}$ was then extrapolated using the Equation~\eqref{eq:max_distance}:

\begin{equation}\label{eq:max_distance}
   N^{}_d = \frac{4^2 N_4^{}}{d^2}
\end{equation}

where $N_d^{}$ is the number of points detected at distance $d$, and $N^{}_4$ is the number of points detected at a distance of 4\,m.

To further enhance the accuracy and reliability of the tracking system, the AST and ADT tracking modalities are fused together. One intuitive method is the naive fusion which ignores the inherent correlations directly. In this sense, this method is too optimistic, and cannot guarantee consistent results \cite{li2023information}.

In our method, we employ two different Kalman filters, each generating its own estimate based on the accumulated data. While these estimates originate from the same LiDAR sensor, the use of separate Kalman filters implies a degree of independence between the two estimation processes. This separation can be seen as a form of distributed processing, where different components (the two Kalman filters) are responsible for generating their own estimates. Its distributed nature can guide the selection of appropriate fusion techniques, such as Inverse Covariance Intersection (ICI)~\cite{noack2017decentralized}, that represents a well-suited fusion rule for typical Kalman filter-based fusion problems \cite{noack2017inverse}.

ICI combines the covariance matrices of the two estimators ($\boldsymbol{\Sigma}_{\text{AST}}$ and $\boldsymbol{\Sigma}_{\text{ADT}}$) using a weighting factor $\omega \in [0,1]$ to generate a fused matrix that more accurately represents the estimated state uncertainty.

The state estimations $\mathbf{x}_{\text{AST}}$ and $\mathbf{x}_{\text{ADT}}$ are fused into a final state $\mathbf{x}_{\text{UAV}}$ using a weighted average:

\begin{equation}\label{eq:fused_estimate}
   \mathbf{x}_{\text{UAV}} = \mathbf{K}_{\text{ICI}} \mathbf{x}_{\text{AST}} + \mathbf{L}_{\text{ICI}} \mathbf{x}_{\text{ADT}}
\end{equation}

where the gains $\mathbf{K}_{\text{ICI}}$ and $\mathbf{L}_{\text{ICI}}$ are computed based on the fused covariance $\boldsymbol{\Sigma}_{\text{UAV}}$ and the individual covariances of the two measurements:

\begin{equation*}
    \boldsymbol{\Sigma}^{}_{\text{UAV}} \hiderel{=} \left(\boldsymbol{\Sigma}_{\text{AST}}^{-1} + \boldsymbol{\Sigma}_{\text{ADT}}^{-1} - \left(\omega \, \boldsymbol{\Sigma}_{\text{AST}} + \left(1-\omega \right) \, \boldsymbol{\Sigma}_{\text{ADT}}\right)^{-1}\right)^{-1}
\end{equation*}

\begin{equation}\label{eq:gains}
    \mathbf{K}_{\text{ICI}} \hiderel{=} \boldsymbol{\Sigma}_{\text{UAV}} \left( \boldsymbol{\Sigma}_{\text{AST}}^{-1} - \omega \left(\omega \, \boldsymbol{\Sigma}_{\text{AST}} + \left(1-\omega \right)\boldsymbol{\Sigma}_{\text{ADT}}\right)^{-1} \right)
\end{equation}

\begin{equation*}
    \mathbf{L}_{\text{ICI}} \hiderel{=} \boldsymbol{\Sigma}_{\text{UAV}} \left( \boldsymbol{\Sigma}_{\text{ADT}}^{-1} - (1-\omega) \left( \omega \, \boldsymbol{\Sigma}_{\text{AST}} + \left(1-\omega\right)\boldsymbol{\Sigma}_{\text{ADT}}\right)^{-1} \right)
\end{equation*}

The optimal value of $\omega$ is found using Brent's algorithm~\cite{brent1971algorithm} to minimize the trace of the inverse covariance matrix:

\begin{dmath}\label{eq:trace}
    \operatorname{trace}\left[ \left( \boldsymbol{\Sigma}_{\text{AST}}^{-1} + \boldsymbol{\Sigma}_{\text{ADT}}^{-1} - (\omega \, \boldsymbol{\Sigma}_{\text{AST}} + (1-\omega) \, \boldsymbol{\Sigma}_{\text{ADT}})^{-1} \right)^{-1} \right]
\end{dmath}
    
By fusing the two estimators at two frequencies, the tracking system can better handle challenging tracking scenarios, such as sensor noise, and provide more accurate and reliable estimates of the UAV state.

\section{Experimental Results}

The experimental platform consist of a Livox Horizon LiDAR  ($81.7\degree \times 25.1\degree$ FoV)  capable of generating non-repetitive point cloud scans up to 100\,Hz, and an external positioning system to validate the extracted trajectories. To test our adaptive tracking algorithm, we evaluated its performance on several trajectories.

\subsection{Metrics}

First, to quantify the disparity between the LiDAR and the external position system estimates, we computed the error by taking the difference between the position estimates obtained from both systems for two distinct positions and orientations of the target. This analysis revealed a Root Mean Squared Error (RMSE) of 0.0143\,m.

To benchmark the proposed approach, we compared two versions of our adaptive method, one using a KF with a CV motion model, and the other using an EKF with a CTRV motion model.  As baseline, we also included tracking keeping the integration time constant. To evaluate the tracking performance quantitatively, we used the RMSE metric, with the main results summarized in Table~\ref{tab:error_adaptive}.

\begingroup
\renewcommand{\arraystretch}{1.5}
\begin{table*}[t]
    \centering
    \caption{Position error (RMSE) for both constant and adaptive integration methods (N/A when the error diverges because the estimated trajectory is incomplete. Unit: meter)}
    \resizebox{0.70\textwidth}{!}{
        \begin{tabular}{@{}l|ccccc|cccccc@{}}
            \toprule
            \textbf{Track} & \multicolumn{11}{c}{\textbf{Tracking Method}} \\
            \hline
             &  $I_{2}^{}$ & $I_{5}^{}$ & $I_{10}^{}$ & $I_{20}^{}$ & $I_{50}^{}$ & $I_{}^{\text{KF}}$ & $I_{\text{AST}}^{\text{KF}}$ & $I_{\text{ADT}}^{\text{KF}}$ & $I_{}^{\text{EKF}}$ & $I_{\text{AST}}^{\text{EKF}}$ & $I_{\text{ADT}}^{\text{EKF}}$\\
            \hline
            T1 & 0.0786 & 0.0774 & 0.0788 & 0.0844 & 0.1019 & 0.0782 & \textbf{0.0773} & 0.0845 & 0.0852 & 0.087 & 0.0861\\
            \hline
            T2 & 0.0202 & \textbf{0.0195} & 0.0227 & 0.0332 & 0.0515 & 0.0253 & 0.0203 & 0.0335 & 0.0395 & 0.0371 & 0.046 \\
            \hline
            T3 & 0.0491 & \textbf{0.0421} & 0.0451 & 0.0553 & 0.0803 & 0.0456 & 0.0431 & 0.0537 & \textbf{0.0421} & 0.0426 & 0.0466 \\
            \hline
            T4 & N/A & N/A & N/A & N/A & N/A & N/A & N/A & N/A & 0.0758 & \textbf{0.0535} & 0.1025 \\
            \hline
            T5 & N/A & 0.0471 & 0.0564 & 0.0841 & 0.1682 & 0.0551 & 0.0501 & 0.0783 & 0.0377 & \textbf{0.0314} & 0.0624 \\
            \hline
            T6 & N/A & 0.0551 & 0.0609 & 0.0915 & N/A & 0.0707 & N/A & 0.0861 & 0.0373 & \textbf{0.0291} & 0.0571 \\
            \hline
            T7 & N/A & N/A & N/A & N/A & N/A & 0.0873 & N/A & N/A & 0.0580 & \textbf{0.0538} & 0.0733 \\
            \hline
            T8 & N/A & N/A & N/A & N/A & N/A & 0.0939 & N/A & 0.1154 & 0.0530 & \textbf{0.0487} & 0.0732 \\
            \hline
            T9 & N/A & N/A & N/A & N/A & 0.2576 & N/A & N/A & N/A & 0.0616 & \textbf{0.0537} & 0.0813 \\
            \hline
            T10 & N/A & 0.0956 & 0.0828 & 0.1089 & 0.1911 & 0.0881 & 0.1167 & 0.1071 & 0.0522 & \textbf{0.0460} & 0.0730 \\
            \bottomrule
        \end{tabular}
        }
    \label{tab:error_adaptive}
\end{table*}

\endgroup

\subsection{Indoor Experiments}

Our results show that the proposed method outperforms the suggested baseline approach. As shown in Table~\ref{tab:error_adaptive}, if the integration time is held constant, the error quickly increases as more scans are integrated. This means that accurate tracking is only possible if the lower end of the range is taken into account.

Despite the inherently higher error with lower scan frequencies, our method offers potential benefits in terms of robustness, efficiency, and flexibility: by adapting the integration time to the UAV motion dynamics, the method is more robust to changes in the environment and can handle unforeseen circumstances, such as sudden changes in direction. Additionally, the adaptive method is more efficient, as it only integrates the number of scans required to obtain accurate state estimates, rather than using a fixed integration time, and  it allows for different integration times to be used in different parts of the trajectory. Therefore, although the combination of the two estimators does not always lead to the best accuracy, at the expenses of a slightly higher RMSE it increases robustness offering potential benefits that could be valuable in certain scenarios: it can effectively fuse the two simultaneous scan frequencies and provide accurate and reliable estimates of the UAV state, even when only one of the two estimators is available. 

As shown in Table~\ref{tab:error_adaptive}, specifically for the tracks T6, T7 and T8, when one or both single estimators fail to track the target, our method can still perform tracking along the whole trajectory by combining both methods using the ICI approach. This demonstrates the effectiveness of our adaptive method in handling challenging tracking scenarios, such as when one estimator is unavailable or unreliable.

Moreover, in 8 out of the 10 analyzed trajectories the EKF demonstrates lower RMSE compared to the KF. Notably, for trajectories T4 and T9, the KF alone fails to effectively complete the tracking task. This suggests that using the CTRV motion model extended to 3D along with an EKF is a viable option also for UAVs.

Furthermore, for both track T1 and T10, although our method does not achieve the overall best performance, it still results in lower errors compared to using the single estimators.

\begin{figure}
    \centering
    \includegraphics[width=0.48\textwidth]{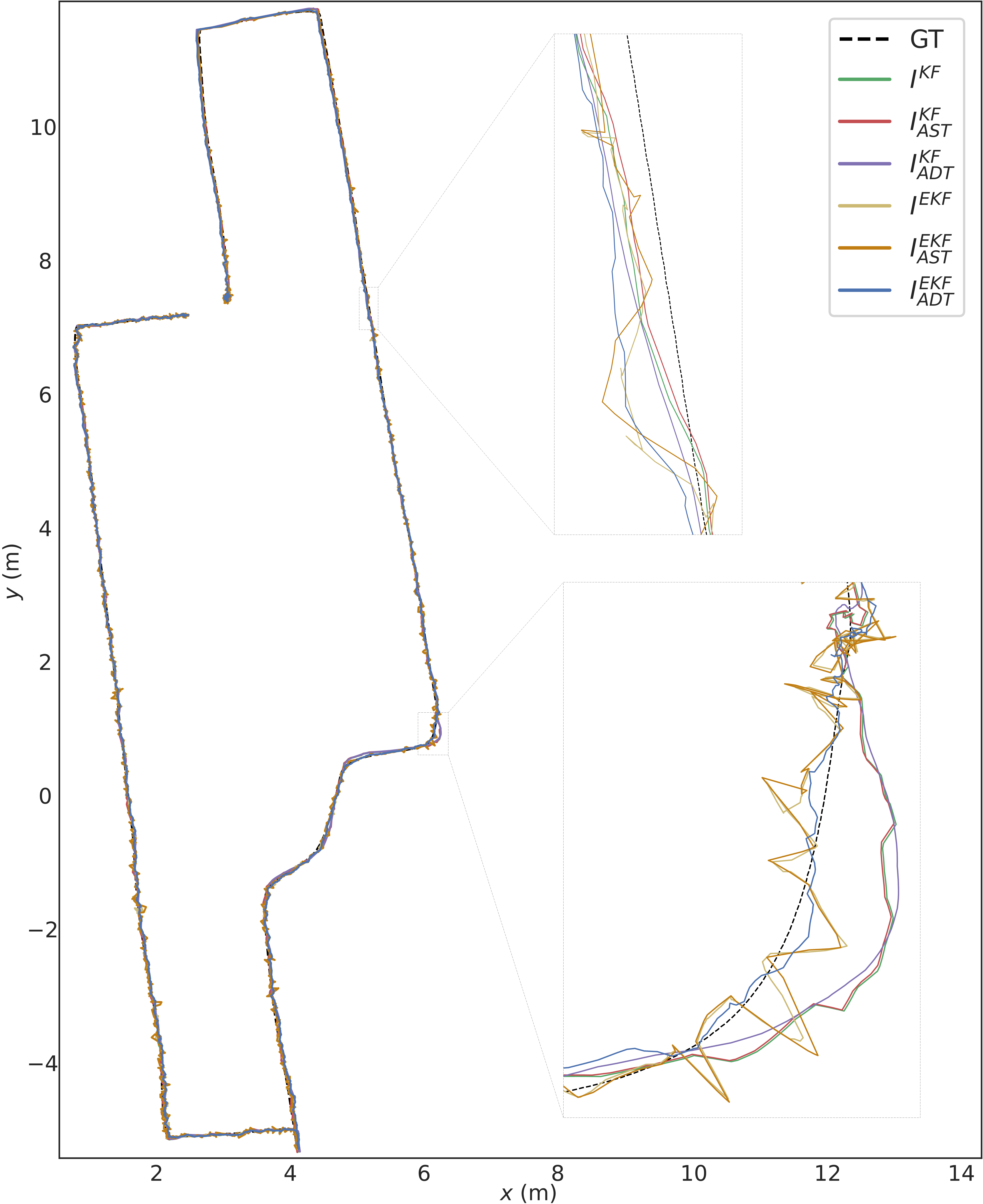}
    \caption{Comparison of trajectories estimated on T1.}
    \label{fig:open_trajectory}
\end{figure}

\begin{figure}
    \centering
    \subfloat[T5]{%
      \includegraphics[width=0.24\textwidth]{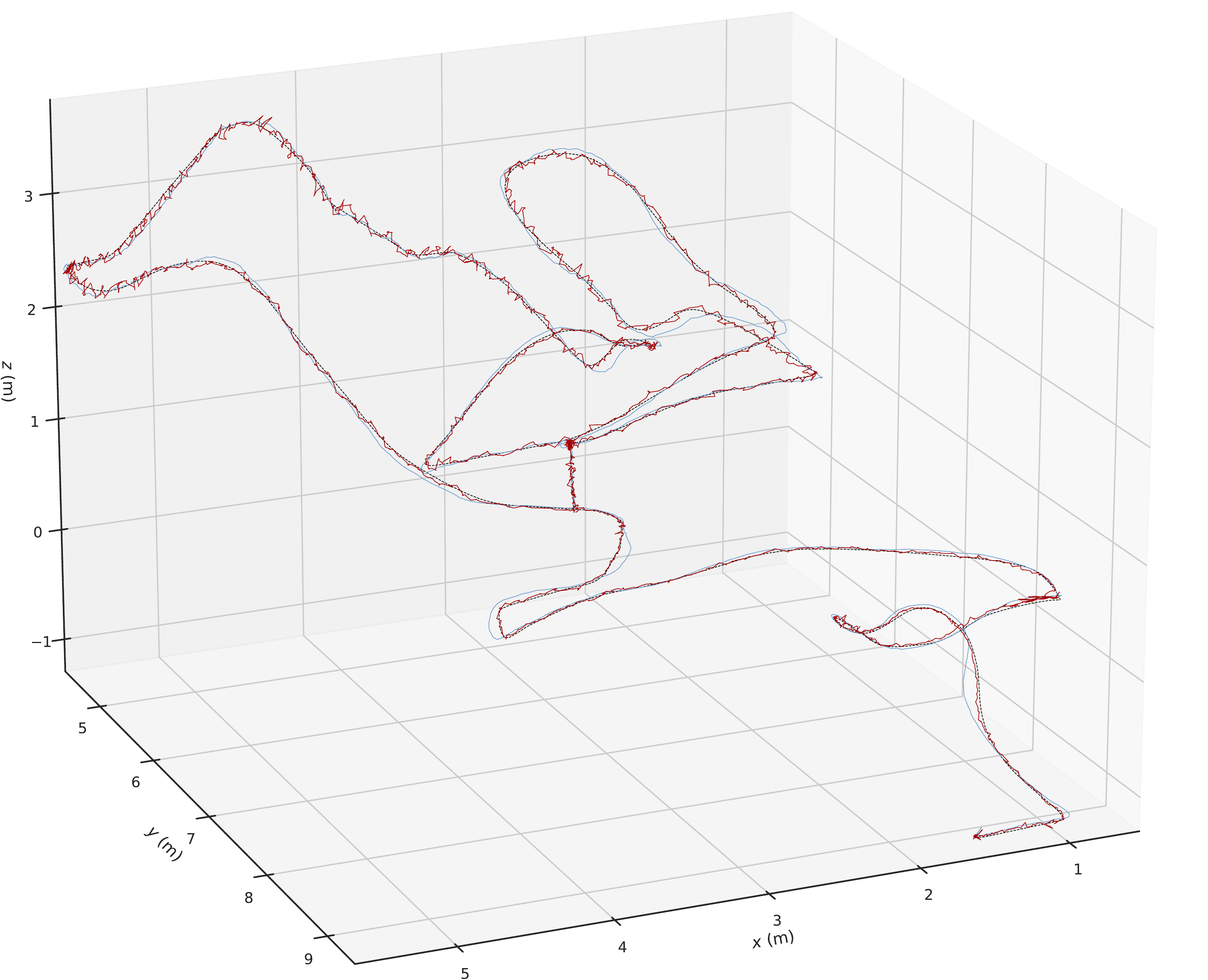}%
      \label{fig:track5}%
    }
    \subfloat[T6]{%
      \includegraphics[width=0.24\textwidth]{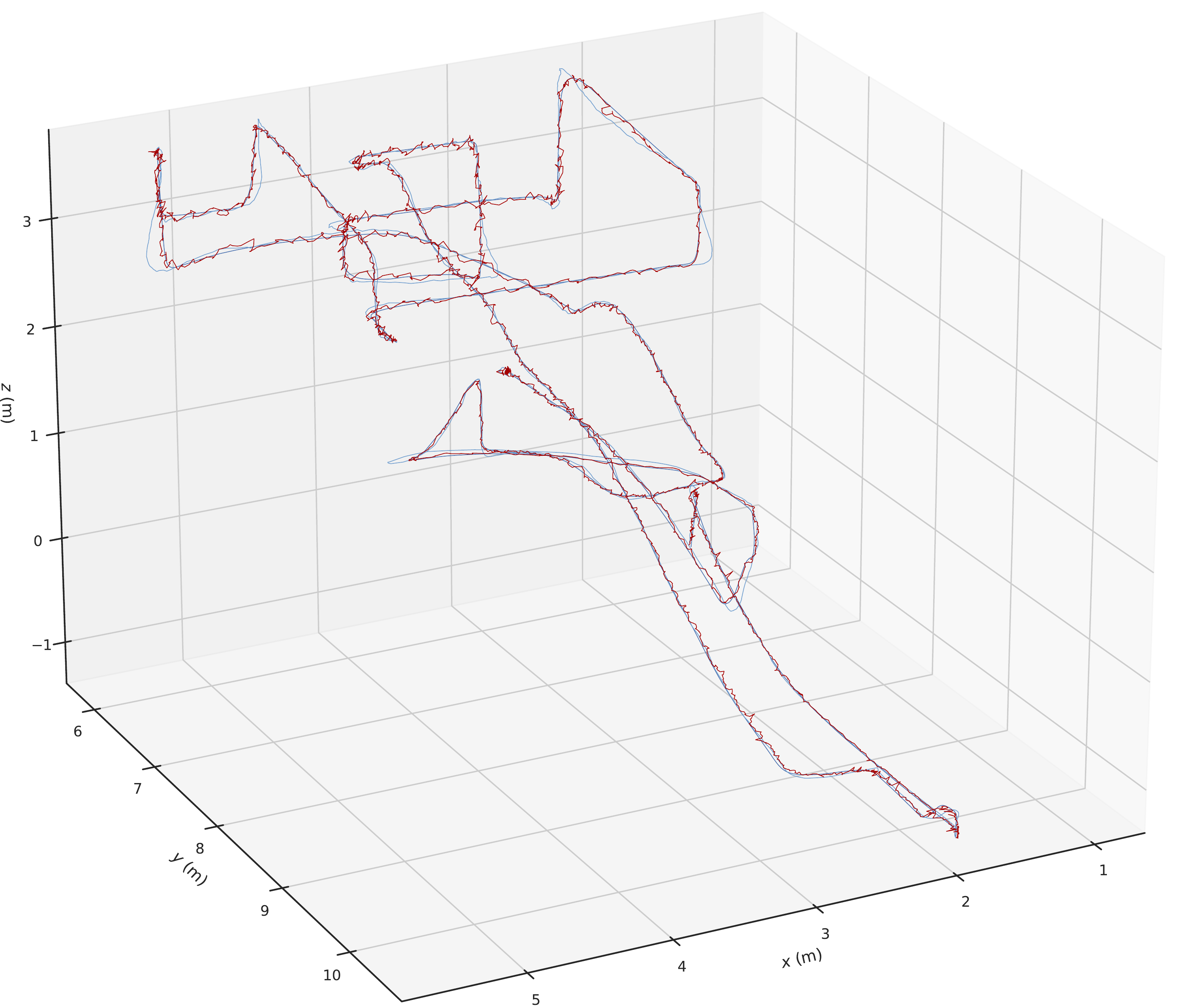}%
      \label{fig:track6}%
    }\qquad
    \subfloat[T8]{%
      \includegraphics[width=0.24\textwidth]{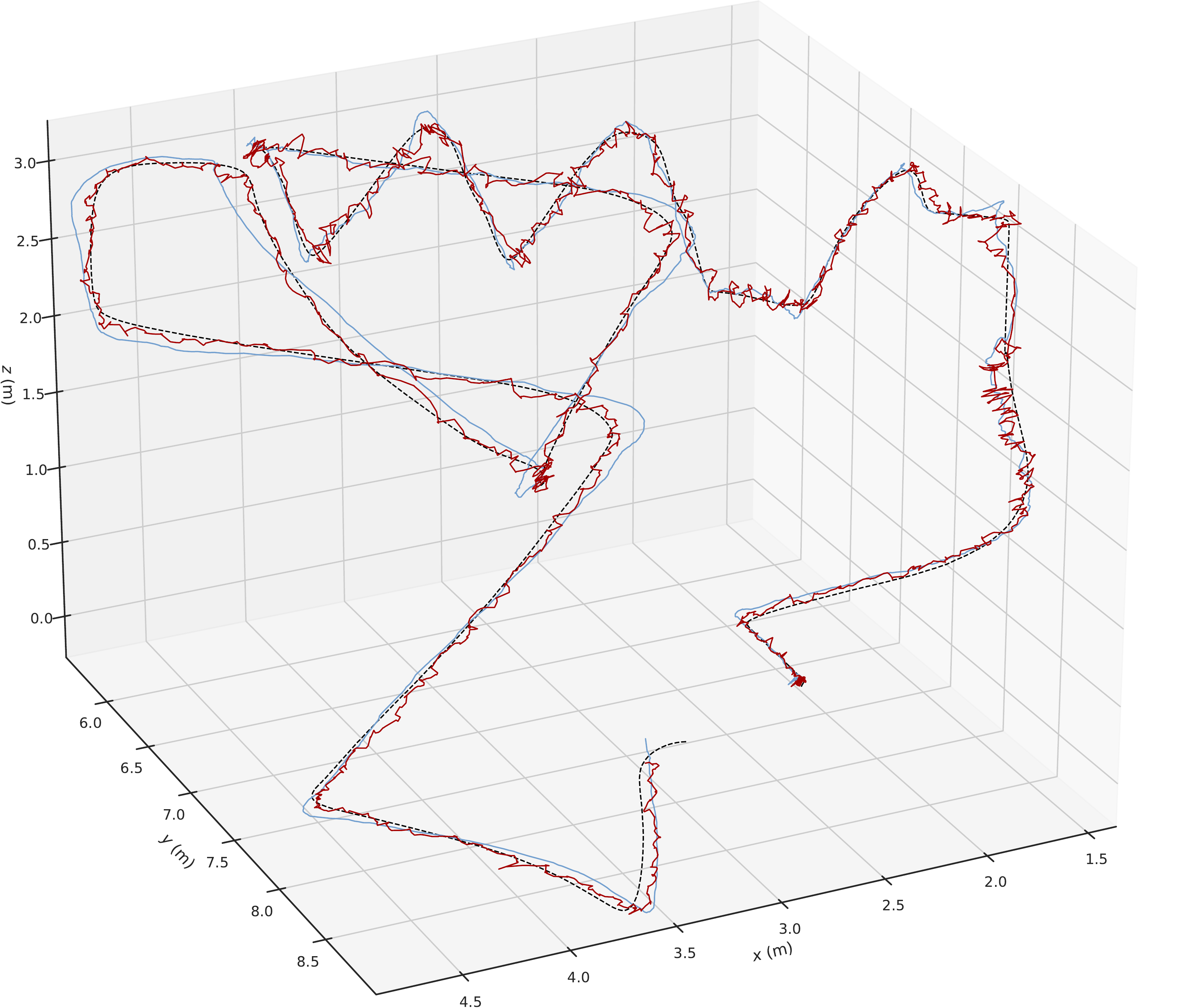}%
      \label{fig:track8}%
    }
    \subfloat[T10]{%
      \includegraphics[width=0.24\textwidth]{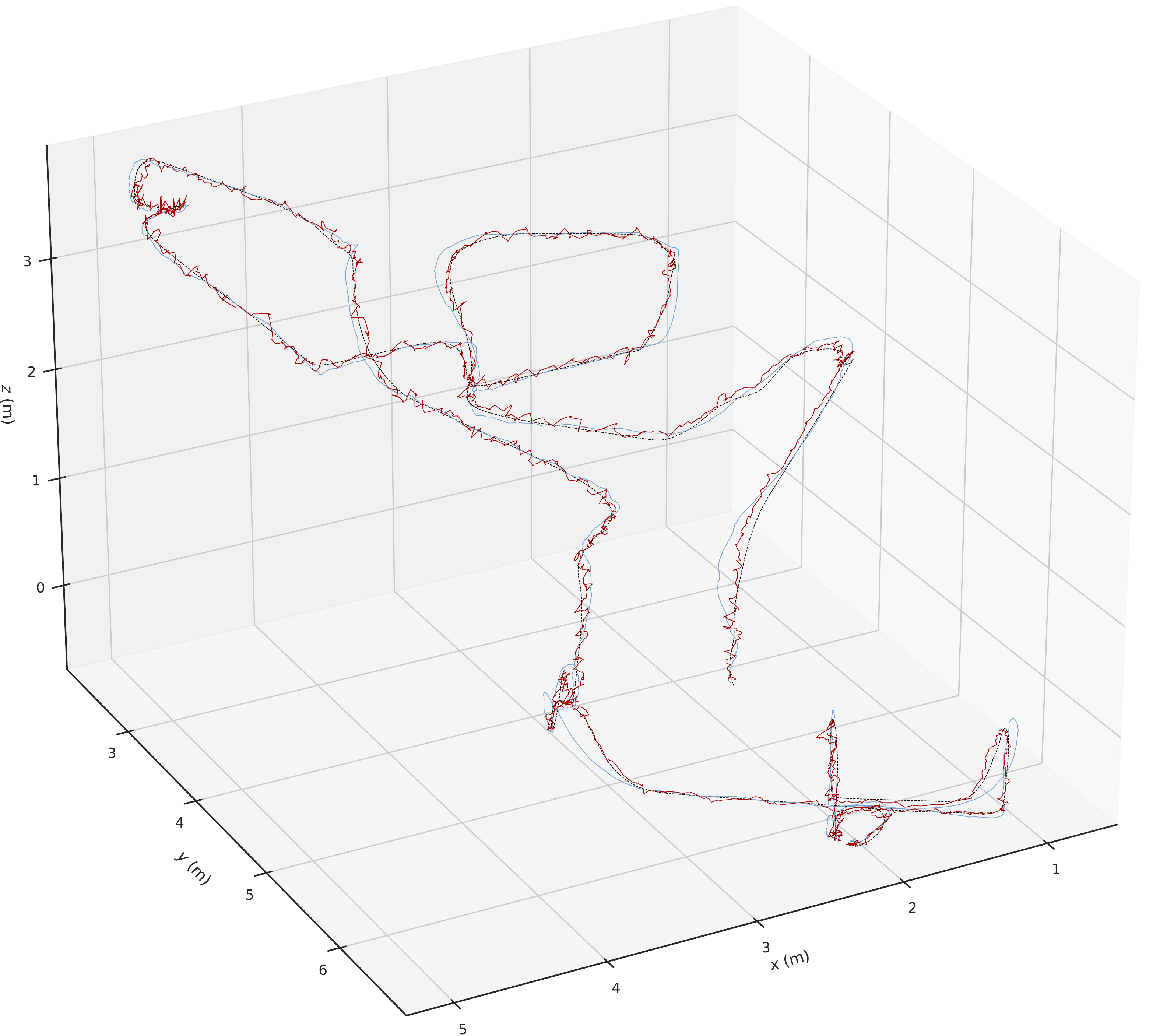}%
      \label{fig:track10}%
    }
    \caption{Visualization of a subset of the trajectories analyzed.}
    \label{fig:estimated_trajectories_sample}
\end{figure}

In addition to the quantitative analysis, we provide visualizations of a subset of the trajectories obtained with our adaptive methods in Fig.~\ref{fig:open_trajectory} and Fig.~\ref{fig:estimated_trajectories_sample}. More results are available on the project's GitHub page. Both methods were able to estimate the overall trajectories well. We also observed that the combination of EKF and CTRV was able to track the target more accurately in situations where the UAV transitioned from a straight motion to a curve, as illustrated in the zoomed portion of Fig.~\ref{fig:open_trajectory}.

\subsection{Initialization and Outdoors Experiments}

While the presented outcomes demonstrate the feasibility of our proposed approach, it is worth noting that the quantitative results rely on the assumption of the initial position being already known. To address this issue, we have developed a method to detect the target's starting location in outdoor scenarios. Drawing inspiration from the effective tracking at short distances demonstrated by Sier et al.~\cite{sier2023uav}, who used signal images generated by a spinning LiDAR, we employed the same custom YOLOv5 model trained on panoramic signal images generated by an Ouster LiDAR~\cite{sier2023uav} to detect the UAV. In our tests, we generate our own range images using a combination of depth and intensity values from a solid-state LiDAR point cloud.

To minimize noise and artifacts when creating a single image, we first integrate a total of 30 frames. Later, the 3D point cloud is projected onto a 2D plane, taking into account both the field of view and image resolution. The transformation process considers both the intensity and distance of each point, combining them through a weighted sum operation to produce the final result. Using only intensity or distance to detect drones from the background becomes challenging for the YOLO model due to the proximity to ground and walls as well as the reflectivity of both the background and the drone being similar.

Furthermore, normalization is applied to ensure appropriate contrast in the resulting image. Upon obtaining the preliminary 2D image, its quality is enhanced through filtering and interpolation: we first identify areas with zero values and substitute them with constants to prevent visual discontinuities. There are two distinct cases that lead to zero-valued pixels after point cloud projection: the sky and other background regions where the emitted laser fails to reflect, and areas within the environment where objects might be present but the LiDAR does not scan. Differentiating between these two cases is crucial for the task of image completion, as it allows for an accurate understanding of the context surrounding the missing pixels.

\begin{figure}
    \centering
    \begin{subfigure}{0.48\textwidth}
        \centering
        \includegraphics[width=\textwidth]{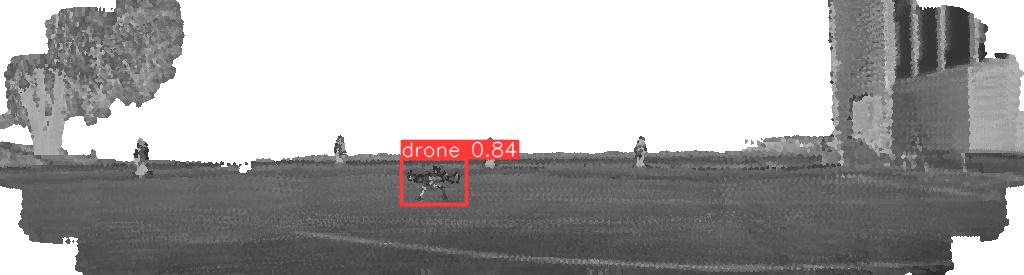}
        \caption{Drone detection on a stationary position on the ground before taking off.}
        \label{fig:detection_ground}
    \end{subfigure}

    \vspace{.42em}
    \begin{subfigure}{0.48\textwidth}
        \centering
        \includegraphics[width=\textwidth]{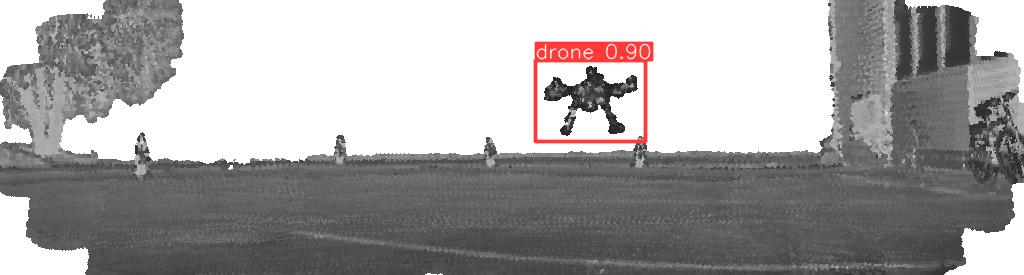}
        \caption{Drone detection with target hovering mid-air.}
        \label{fig:detection_air}
    \end{subfigure}

    \vspace{.42em}
    \begin{subfigure}{0.48\textwidth}
        \centering
        \includegraphics[width=\textwidth]{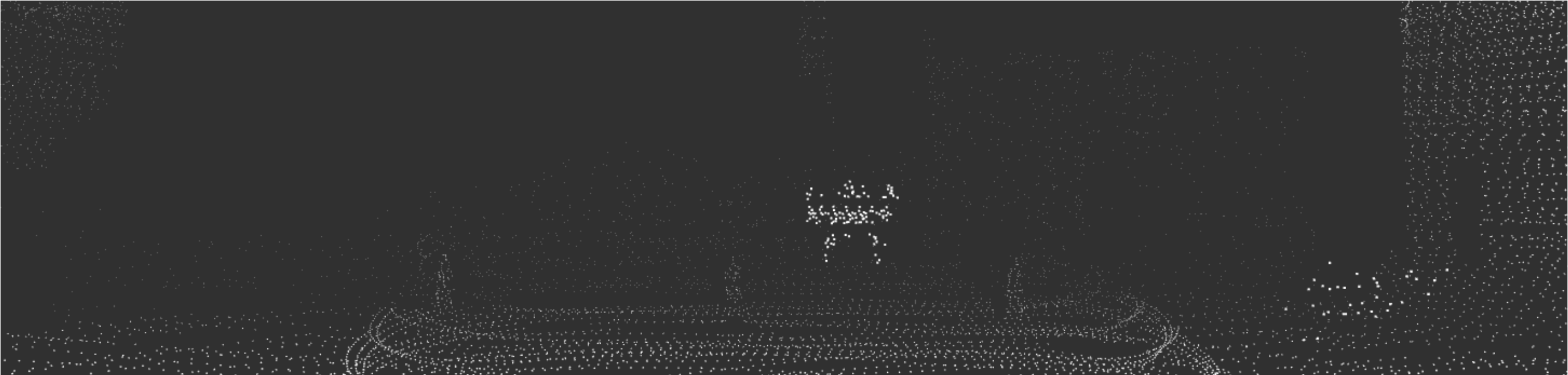}
        \caption{Raw point cloud with target hovering mid-air (0.3\,s integration).}
        \label{fig:detection_air_pointcloud}
    \end{subfigure}
    \caption{Results of the YOLOv5 object detector trained in~\cite{sier2023uav}, applied on the generated range images in the outdoors experiment (a)-(b). In (c), we show a sample point cloud with the target hovering in mid-air.}
    \vspace{1em}
    \label{fig:yolo_detection}
\end{figure}

To remove noise artifacts, we use binary thresholding and a nearest-neighbor interpolation to fill in missing or noisy regions, which results in smoother and more accurate images. Figure~\ref{fig:yolo_detection} showcases the final generated depth map and the result of YOLOv5 object detection: the model was able to accurately detect the UAV and determine its initial position based on these images with no need for further training. 

\begin{figure}
    \centering
    \includegraphics[width=0.48\textwidth]{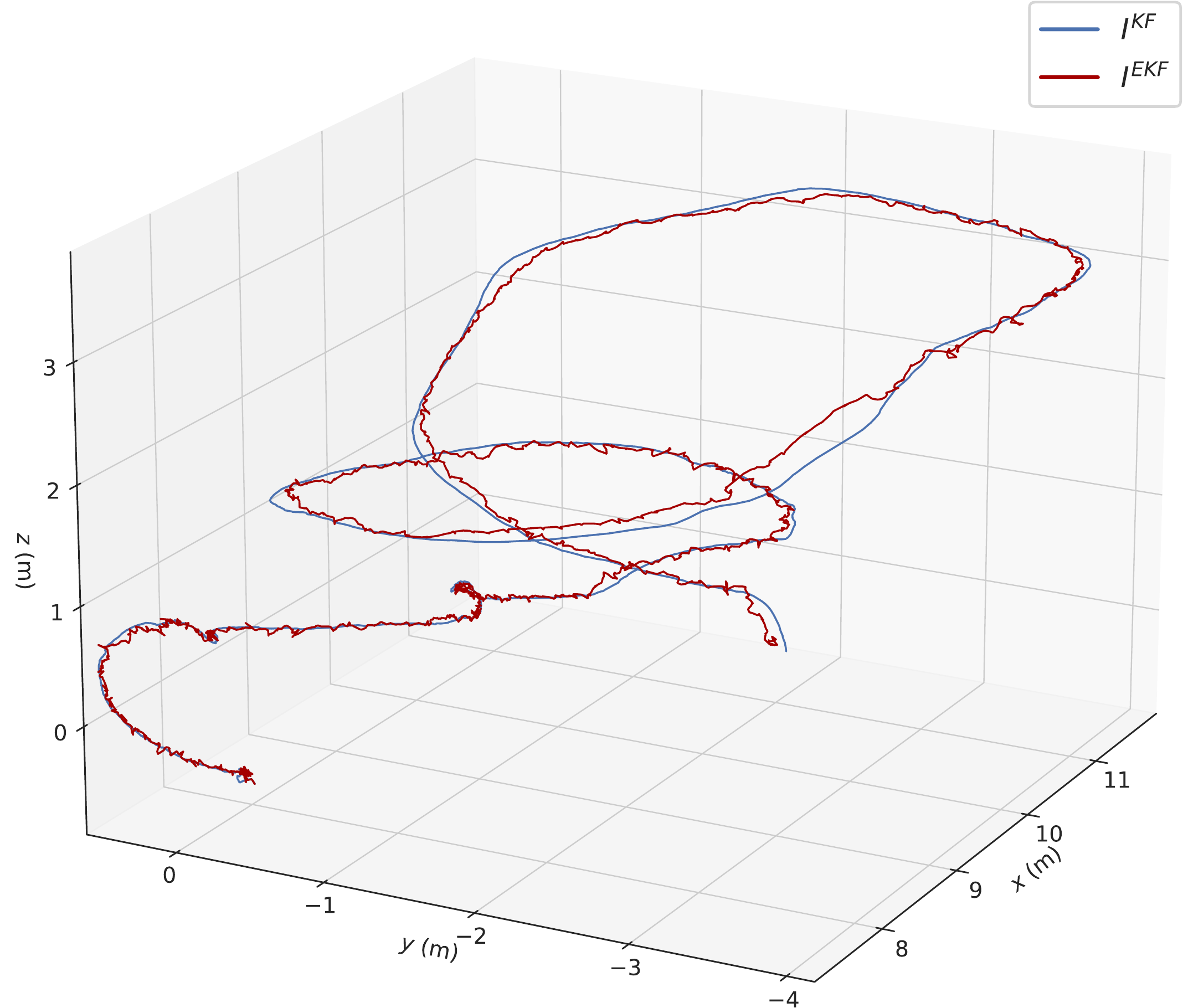}
    \caption{Trajectories estimated with the adaptive method and initialization from range images.}
    \label{fig:outdoor_trajectory}
\end{figure}

Finally, in Fig.~\ref{fig:outdoor_trajectory} we provide a visualization of the tracking algorithm performed with our adaptive method and unknown initial position. Notably, the $z=0$ line in the figure corresponds to the ground instead of the LiDAR reference line. It should be noted that, due to the outdoor nature of the recorded data, ground truth data is not available.

\subsection{Ablation study}

To assess the impact of different weighting functions on the performance of our proposed method, we conducted an ablation study. Specifically, we computed the root mean square error (RMSE) for each track using three different configurations:

\begin{enumerate}[label=(\roman*)]
    \item No weighting function: this configuration assigns equal weight to all points in the track, without considering their temporal order.
    \item Linear decay: the weight assigned to each point decreases linearly with time.
    \item Exponential decay: the weighting function is described in Equation~\eqref{eq:points_weight} in Section III.
\end{enumerate}

By systematically evaluating the effects of different weighting functions, we have validated the effectiveness of our choice, as presented in Table~\ref{tab:error_ablation_weight}.

\begingroup
\renewcommand{\arraystretch}{1.5}
\begin{table*}[t]
    \centering
    \caption{Position error (RMSE) for our adaptive integration method for different types of weight functions (N/A when the error diverges because the estimated trajectory is incomplete. Unit: meter)}
    \resizebox{0.98\textwidth}{!}{
        \begin{tabular}{@{}l|cccccc|cccccc|cccccc@{}}
            \toprule
            \textbf{Track} & \multicolumn{6}{c|}{\textbf{No Decay}} & \multicolumn{6}{c|}{\textbf{Linear Decay}} & \multicolumn{6}{c}{\textbf{Exponential Decay}}\\
            \hline
             &  $I_{}^{\text{KF}}$ & $I_{\text{AST}}^{\text{KF}}$ & $I_{\text{ADT}}^{\text{KF}}$ & $I_{}^{\text{EKF}}$ & $I_{\text{AST}}^{\text{EKF}}$ & $I_{\text{ADT}}^{\text{EKF}}$ & $I_{}^{\text{KF}}$ & $I_{\text{AST}}^{\text{KF}}$ & $I_{\text{ADT}}^{\text{KF}}$ & $I_{}^{\text{EKF}}$ & $I_{\text{AST}}^{\text{EKF}}$ & $I_{\text{ADT}}^{\text{EKF}}$ & $I_{}^{\text{KF}}$ & $I_{\text{AST}}^{\text{KF}}$ & $I_{\text{ADT}}^{\text{KF}}$ & $I_{}^{\text{EKF}}$ & $I_{\text{AST}}^{\text{EKF}}$ & $I_{\text{ADT}}^{\text{EKF}}$\\
            \hline
            T1 & 0.0788 & \textbf{0.0773} & 0.0882 & 0.08615 & 0.087 & 0.0901 & 0.0814 & 0.0783 & 0.0973 & 0.0863 & 0.0886 & 0.101 & 0.0782 &\textbf{0.0773} & 0.0845 & 0.0852 & 0.087 & 0.0861\\
            \hline
            T2 & 0.023 & 0.0197 & 0.0346 & 0.0375 & 0.0365 & 0.0471 & 0.0309 & \textbf{0.0195} & 0.0439 & 0.04062 & 0.0371 & 0.0565 & 0.0253 & 0.0203 & 0.0335 & 0.0395 & 0.0371 & 0.046 \\
            \hline
            T3 & N/A & N/A & N/A & N/A & N/A & N/A & 0.0506 & 0.0438 & 0.0661 & 0.0451 & 0.0444 & 0.0601 & 0.0456 & 0.0431 & 0.0537 & \textbf{0.0421} & 0.0426 & 0.0466 \\
            \hline
            T4 & N/A & N/A & N/A & 0.7311 & 0.728 & 0.7409 & N/A & N/A & N/A & 0.735 & 0.7298 & 0.7475 & N/A & N/A & N/A & 0.0758 & \textbf{0.0535} & 0.1025 \\
            \hline
            T5 & 0.0585 & N/A & 0.0872 & 0.0432 & \textbf{0.0314} & 0.0722 & 0.0674 & N/A & 0.1053 & 0.0503 & 0.0331 & 0.0937 & 0.0551 & 0.0501 & 0.0783 & 0.0377 & \textbf{0.0314} & 0.0624 \\
            \hline
            T6 & 0.0708 & N/A & 0.0907 & 0.0393 & \textbf{0.0288} & 0.067 & 0.0778 & N/A & 0.1044 & 0.0467 & 0.0311 & 0.0837 & 0.0707 & N/A & 0.0861 & 0.0373 & 0.0291 & 0.0571 \\
            \hline
            T7 & N/A & N/A & N/A & 0.06 & \textbf{0.0538} & 0.0797 & 0.09 & N/A & N/A & 0.0645 & 0.0551 & 0.0933 & 0.0873 & N/A & N/A & 0.058 & \textbf{0.0538} & 0.0733 \\
            \hline
            T8 & N/A & N/A & N/A & N/A & N/A & N/A & 0.38 & N/A & N/A & N/A & N/A & N/A & 0.0939 & N/A & 0.1154 & 0.053 & \textbf{0.0487} & 0.0732 \\
            \hline
            T9 & N/A & N/A & N/A & 0.0629 & \textbf{0.0534} & 0.0882 & N/A & N/A & N/A & 0.0677 & 0.0542 & 0.1058 & N/A & N/A & N/A & 0.0616 & 0.0537 & 0.0813 \\
            \hline
            T10 & 0.1073 & 0.1153 & 0.1136 & 0.0539 & 0.0461 & 0.0955 & N/A & 0.1178 & 0.1303 & 0.0605 & 0.0475 & 0.1028 & 0.0881 & 0.1167 & 0.1071 & 0.0522 & \textbf{0.0460} & 0.073 \\
            \bottomrule
        \end{tabular}
    }
    \label{tab:error_ablation_weight}
\end{table*}

\endgroup

The usage of a weight function improves robustness (specifically on tracks T3, T7, T8). In particular, using an exponential weighting function almost always (except for track T2 and T4) reduces the RMSE for the combined estimation and the estimation obtained accumulating a higher number of scans ($I^{\text{EKF}}_{\text{ADT}}$).

Similarly, we conducted ablation studies to verify the effectiveness of the ICI as fusion strategy. We compared it against a simple and a weighted average, where the covariance matrices of each estimator was used to weight its contribution. As shown in Table~\ref{tab:error_ablation_ici}, for 7 out of the 10 tracks analyzed, ICI reports a lower RMSE.

\begingroup
\renewcommand{\arraystretch}{1.5}
\begin{table}[t]
    \centering
    \caption{Position error (RMSE) for our adaptive integration method for different types of fusion strategies (N/A when the error diverges because the estimated trajectory is incomplete. Unit: meter)}
    \resizebox{0.48\textwidth}{!}{
    \begin{tabular}{@{}l|cc|cc|cc@{}}
        \toprule
        \textbf{Track} & \multicolumn{2}{c|}{\textbf{Simple Average}} & \multicolumn{2}{c|}{\textbf{Weighted Average}} & \multicolumn{2}{c}{\textbf{ICI}}\\
        \hline
         &  $I_{}^{\text{KF}}$ & $I_{}^{\text{EKF}}$ & $I_{}^{\text{KF}}$ & $I_{}^{\text{EKF}}$ & $I_{}^{\text{KF}}$ & $I_{}^{\text{EKF}}$\\
        \hline
        T1 & 0.0795 & 0.0839 & 0.0799 & 0.0867 & \textbf{0.0782} & 0.0852\\
        \hline
        T2 & \textbf{0.0235} & 0.0361 & 0.0236 & 0.0361 & 0.0253 & 0.0395\\
        \hline
        T3 & 0.0472 & \textbf{0.041} & 0.0479 & 0.0413 & 0.0456 & 0.0421\\
        \hline
        T4 & N/A & \textbf{0.733} & N/A & N/A & N/A & 0.0758\\
        \hline
        T5 & 0.0617 & 0.0429 & 0.0634 & 0.044 & 0.0551 & \textbf{0.0377}\\
        \hline
        T6 & N/A & 0.0395 & N/A & 0.0409 & 0.0707 & \textbf{0.0373}\\
        \hline
        T7 & N/A & 0.0607 & N/A & 0.0618 & 0.0873 & \textbf{0.058}\\
        \hline
        T8 & N/A & N/A & N/A & N/A & 0.0939 & \textbf{0.053}\\
        \hline
        T9 & N/A & 0.0650 & N/A & 0.0667 & N/A & \textbf{0.0616}\\
        \hline
        T10 & 0.0997 & 0.0554 & 0.0929 & 0.0565 & 0.0881 & \textbf{0.0522}\\
        \bottomrule
    \end{tabular}
    }
    \label{tab:error_ablation_ici}
\end{table}

\endgroup

\section{Conclusion}\label{sec:conclusion}

In this paper, we presented a novel adaptive tracking approach for UAVs that fuses tracking information from two different scan frequencies from a single solid-state LiDAR sensor. One of the frequencies allows for high accuracy while the second enables more persistent tracking. Our method dynamically adjusts the LiDAR's frame integration time based on the UAV's travelling speed and distance from the sensor, allowing for accurate estimates of the UAV's state.

The experimental results demonstrate that by tailoring the frame integration time to the UAV's movement characteristics, our method outperforms the established baseline method, while also providing more reliable and precise tracking when the estimator from one of the scan frequencies is unavailable or unreliable. The proposed method leverages the Inverse Covariance Intersection method and Kalman filters to enhance tracking accuracy and handle challenging tracking scenarios, suggesting that using the CTRV motion model extended to 3D along with an EKF is a viable option also for UAVs.

In addition, to overcome the challenge of needing a known initial position for detection, we have developed a solution that generates range images in outdoor environments. These range images, created by combining depth and intensity data, can then be used by a YOLOv5 model to accurately detect the UAV's initial position.

In future works, we plan to explore the integration of LiDAR-based tracking into UAV navigation and the integration of other external sensors to further improve the accuracy and robustness of our method.


\section*{Acknowledgment}

This research work is supported by the Academy of Finland's Aeropolis project (Grant No. 348480).

\bibliographystyle{unsrt}
\bibliography{shortbib}

\end{document}